# Social network analysis for personalized characterization and risk assessment of alcohol use disorders in adolescents using semantic technologies

José Alberto Benítez-Andrades [a,*], Isaías García-Rodríguez [b], Carmen Benavides [a], Héctor Alaiz-Moretón [b], Alejandro Rodríguez-González [c,d]

[a] SALBIS Research Group, Department of Electric, Systems and Automatics Engineering, University of León, Campus of Vegazana s/n, León, 24071, León, Spain
[b] SECOMUCI Research Group, Escuela de Ingenierías Industrial e Informática, Universidad de León, Campus de Vegazana s/n, C.P. 24071 León, Spain
[c] Centro de Tecnología Biomédica, Universidad Politécnica de Madrid, Spain
[d] Escuela Técnica Superior de Ingenieros Informáticos, Universidad Politécnica de Madrid, Spain

ABSTRACT

Alcohol Use Disorder (AUD) is a major concern for public health organizations worldwide, especially as regards the adolescent population. The consumption of alcohol in adolescents is known to be influenced by seeing friends and even parents drinking alcohol. Building on this fact, a number of studies into alcohol consumption among adolescents have made use of Social Network Analysis (SNA) techniques to study the different social networks (peers, friends, family, etc.) with whom the adolescent is involved. These kinds of studies need an initial phase of data gathering by means of questionnaires and a subsequent analysis phase using the SNA techniques. The process involves a number of manual data handling stages that are time consuming and error-prone. The use of knowledge engineering techniques (including the construction of a domain ontology) to represent the information, allows the automation of all the activities, from the initial data collection to the results of the SNA study. This paper shows how a knowledge model is constructed, and compares the results obtained using the traditional method with this, fully automated model, detailing the main advantages of the latter. In the case of the SNA analysis, the validity of the results obtained with the knowledge engineering approach are compared to those obtained manually using the UCINET, Cytoscape, Pajek and Gephi to test the accuracy of the knowledge model.

## 1. Introduction

Adolescence is a stage of life in which the individual is evolving towards adulthood. This implies a great number of changes both at the physical and emotional levels, until he or she reaches the status of independence that defines the adult individual [1]. In this context, aspects related to the socialization and identity of the individual play a determinant role. Specifically, not only are significant relationships established with the group of peers or friends, but also those taking place in the family environment in which the individual is growing and acquiring the significant values for adulthood [2,3].

* Corresponding author.
E-mail addresses: jbena@unileon.es (J.A. Benítez-Andrades),
igarr@unileon.es (I. García-Rodríguez), carmen.benavides@unileon.es
(C. Benavides), hector.moreton@unileon.es (H. Alaiz-Moretón),
alejandro.rg@upm.es (A. Rodríguez-González).

Alcohol consumption has been identified as one of the most important public health problems worldwide, not only leading to many of the use disorders and dependence problems it causes in the long term, but also because of the direct relationship that has been proven in traffic accidents, inappropriate sexual behavior or confusing psychological perspectives [4]. Alcohol is usually the most consumed drug among Spanish adolescents [5], establishing it as normal behavior within this age range [6]. It is precisely because of this normalization in drinking alcohol that there is a great political and social concern as regards alcohol consumption among adolescents [7].

Traditionally, research and practice into alcohol use disorder risk have focused on the individual, with the assumption that lifestyle and health behaviors are decided primarily by individual choice. But many detailed studies have proven that the etiology of adolescent alcohol (and other drug) use is complex and can be studied at different levels. A number of works have studied the extent to which the social relationships of the adolescent influence his/her alcohol consumption. These studies include some, or all, of the family, school, friend and peer social environments and have shown that relationships established within these social networks do have an effect on alcohol consumption habits. A good review of some of these studies for the European Union is Steketee et al. [8]. More detailed studies, focusing on classroom environments, have used social network analysis (SNA) and sophisticated stochastic actor-based models on the influence and selection of networks [9]. The study detailed in Osgood et al. [10], was carried out on a large adolescent population and across data collection campaigns and analysis stages; this study found clear evidence on both selection and influence dynamics regarding alcohol use habits (selection refers to the fact that boys and girls with similar levels of alcohol consumption tend to join together to form relationships, while influence is the process through which the consumption of alcohol in a group may change that of a new individual who joins the group). These influence mechanisms have also been observed in other substance consumption habits such as tobacco [11] or cannabis [12] and even social behavior such as delinquency [13].

The systematic exploitation of the complex information involving both the social relationships and personal circumstances of the adolescent may facilitate the understanding of the pattern of his or her present and future alcohol use. Social Network Analysis (SNA) is a discipline dedicated to studying the structure and the dynamics of the interactions in a social network, so its methods can be used for this purpose [14–17]. SNA defines and uses different metrics, methods and algorithms to be applied to a network of relationships in order to gain an insight into its structure and dynamics, but it requires specific knowledge of the field and the use of specialized software tools in order to apply the techniques and interpret the results. The processes involved in an SNA study require time-consuming and error-prone procedures, as many of them are usually accomplished manually. Therefore, the problem to be solved in this research is to perform analysis of social networks to different networks automatically and also to draw conclusions to that analysis of social networks, avoiding possible manual errors in the manipulation of data, saving time in its preparation and in its final result.

The objective of this research is to create a computer tool that can be used by any professional from the health sector who is interested in gaining an insight into the patterns of alcohol consumption and the personal situations of a group of people involved in alcohol use, in particular in adolescent populations. The tool should apply SNA tools and methods conveniently but hide the details to the final user, in such a way that the health professional does not have to know anything about them, only the results of the analysis will be revealed, consisting of terms and graphics which are familiar to him/her. Moreover, the tool must allow the automation of all the stages of the study, from the data gathering to the drawing up of conclusions, with no data handling involving humans.

In order to achieve these objectives, techniques from semantic knowledge representation have been applied to build the application described. These techniques allow the construction of conceptual structures, called ontologies [18], that capture and connect the knowledge of the different domains involved (personal data, relationships, alcohol consumption terms, SNA techniques, psychosocial health care terminology, etc.) [19,20] and allow the computer to handle them at a high level of abstraction. This kind of development needs the close collaboration of psychosocial healthcare professionals with knowledge of alcohol use disorders, experts from social network analysis domains, knowledge engineers and graphical user interface designers and programmers.

The tool was developed for use by researchers and practitioners from the healthcare sector whose objective is the study of the role that social relationships play in patterns of alcohol consumption in adolescents, or just to get an insight into (or monitor) the current relationships among a group of students, including their alcohol consumption habits.

In short, it is possible to say that this research has led to the following contributions that will be explained throughout this manuscript:

- Developing of a knowledge base, based on a multi-domain ontology including classes that represent different concepts about people, questionnaires and a social network analysis.
- Developing of a knowledge-based application based on this conceptual model.
- Validation of the model by comparing the results obtained for the social network analysis metrics with other existing validated tools.
- Building a web application that is able to gather the data using questionnaires and by analyzing the information obtained automatically without any human intervention or need for manual data handling, using the ontology and the knowledge base.

The rest of the paper is organized as follows. Section 2 introduces Social Network Analysis and how it can help healthcare studies, with an emphasis on adolescent social networks and alcohol consumption; it also describes the usual process for applying SNA to healthcare studies into alcohol consumption. Section 3 introduces the domain of semantic knowledge representation in the computer, and experience in the application of these techniques in SNA studies. Section 4 describes the research, presenting the requirements for the application to be built, the ontologies developed and the application programmed. Section 5 shows an experimental test achieved in a real scenario and includes information about the validation of the tool. Finally, Section 6 gives some discussion and conclusions.

## 2. Social network analysis and its application to the study of alcohol consumption habits

The objective of Social Network Analysis (SNA) is to find and exploit the most important features of social networks with the aim of finding out how network nodes (usually representing people) are connected, and how these relationships characterize the flow of information within the network. It also helps to analyze the different groups or communities that can be detected and the individuals that play, or can play, important roles within a group, in the whole network, or as mediators between nodes or groups. SNA techniques are also used for studying and predicting the evolution of social networks. For these reasons, SNA has been used in organizations and companies as an important tool for knowing and managing their human capital. Next, a brief introduction to SNA will be presented, showing the main metrics, their meanings and use. Section 2.2 is devoted to detailing how SNA techniques have been applied to the study of alcohol consumption habits, in order to show the suitability of the approach. Section 2.3 shows how an SNA study into alcohol consumption is usually carried out. This introduction will serve as a starting point to show how the semantic knowledge representation can facilitate this kind of study.

### 2.1. A brief introduction to social network analysis

A social network is comprised of *nodes* (also known as vertices, actors or agents) connected by *edges* (also known as arcs or links) representing the relationships established between nodes. Edges may have directionality, that is, go from one given source node to an end node (the link leaves the origin node and enters the destination node) or may be undirected (where the edge can be considered in either direction). Edges may have a weight which is a value representing the importance of this edge, or not. The set of nodes and edges is called a *graph*. Most simple graphs have only one type of node. Graphs in which there are two types of node are called bipartite graphs, if there are three types of node they are called tripartite graphs and, in general, the term 'multipartite graphs' is used for graphs with nodes having more than three types of node.

Within a *graph*, a *path* is a list of nodes, each of them linked to the next one by an edge. A *directed path* is a sequence of directed edges from a source node to an end node. A *complete graph* would be a graph having an edge between any two pairs of nodes, while a *connected graph* is one having a path between any two pairs of its nodes.

Traditionally SNA has used graph theory for studying the social networks [21,22]. Metrics used by SNA to characterize a network and its components can be grouped into two categories:

- Metrics that provide information on the overall structure and overall disposition of nodes and edges in the network.
- Metrics that provide insight into the most relevant actors involved in the different social relationships and the way they are connected and communicate with other actors and groups.

*Metrics for getting an insight into the overall structure of the network*

The *density* gives an idea of the cohesion of the network, that is, the number of connections that actually exist in relation to the number of total possible connections. This measurement can be used in an egocentric context (measuring density around the node that is the object of the study) or in a socio-centric context (the measurement is taken over the whole graph) [21].

A *geodesic* is the shortest sequence of actors/edges that lies in the shortest path between two given nodes.

The *diameter* is the length of the longest geodesics of the network. The degree distribution follows a power law, a small number of actors has a high degree while many have a low one. As was stated when Milgram studied the "small world effect" [23] the diameter in a social network having n actors is of order log(n).

*Communities* [21] are groups of actors having a dense structure of links between them. Social networks (especially those in which actors involve humans) tend to establish more or less isolated groups of nodes due to the clustering tendency consisting of the predisposition to establish links between others with similar characteristics. The size of the largest component serves as an indicator of the efficiency of the network for communication processes. There are many community detection algorithms that try to find these groupings by using subtly different criteria. There are two main categories for these algorithms: those based on a hierarchical breakdown and those based on heuristics [24].

The number and characteristics of the different communities, as well as the way they are connected to others determine the way actors are distributed in the network and how information flows; it is also significant in the way actors behave, influence others, or are influenced by others [25,26]. Several types of group can be found in a social network. Scott [21] proposed three graph patterns corresponding to three types of group one can find in a social network: *components* (isolated connected sub-graphs), *cliques* (complete sub-graphs), and *cycles* (paths returning to their point of departure). These patterns were extended in order to detect finer-grain communities: An *n-clique* is a clique of actors having a maximum distance of n to any other member of the group; a *k-plex* is defined as a group in which each member is connected to all the other members except for a maximum of k actors. A *triad* is a cycle with a length of 3 (it connects three different nodes).

*Metrics for information about individual actors and strategic positions*

*Centrality measures* focus on individual actors (nodes) in the network. They try to characterize the actors and find those who play important, structural, roles in the social network. Three main types of centrality can be calculated [27]: degree centrality, closeness centrality and betweenness centrality.

The *degree centrality* considers the number of edges attached to a node. Nodes with higher degree centrality are considered as more central, very visible, and as local actors influencing their neighborhood. In directed graphs the in-degree (number of edges entering the node) and out-degree (number of edges leaving the node) are considered as indicators of the support and influence of the actor, respectively [28]. The *n-degree* is a measure that considers the neighborhood at a distance of n or less actors [29,30], the n-degree of an actor is the number of other actors that this actor is linked to by a sequence of n or less edges. In practice a number of n greater than 2 is not considered as it has been shown that beyond this reach, neither visibility nor influence are relevant [26]. The 2-degree of an actor is called the *reach* of the actor. *Closeness centrality* is obtained with the average length of the paths (number of links) linking a node to others, it measures the capacity of a node of being reached or reaching other actors; if the edges have directionality, the interpretation of the closeness centrality is differentiated between the capacity to reach on one side and the capacity of being reached on the other. *Betweenness centrality* measures the capacity of an actor of being an intermediary between any two others [26], actors with a high betweenness centrality play an important role in the network because of their strategic position [26,31]. The computation of this measure is time consuming and, for big networks, approximating algorithms have been developed. An actor located in a geodesic path has an important position in the cohesion of a network and the flow of information. If actors with a high betweenness centrality are removed from the network, its structure and characteristics could be severely affected. If edges have direction, only paths without any change in direction must be considered.

## 2.2. Social network analysis applied to the study of alcohol consumption habits

Social network analysis tools and techniques have found an increasing number of applications in the healthcare domain in recent years [32]. These applications have focused on aspects such as service provision and organization, studying behavior change (including the diffusion of innovations, opinion leaders and other aspects of social influence), decision-making processes, interpersonal relations, information sharing or social support, among others [33].

As regards alcohol consumption studies, SNA has been used for a number of issues. Studying the mechanism by which the alcohol consumption level of a given individual changes is one of the more demanding and challenging topics. A considerable amount of research has already highlighted the existence of an influence mechanism regarding alcohol consumption in networks of adolescents [34–36] . Since the development of stochastic actor-oriented models [37,38] for studying the dynamics of networks, and the use of social network analysis techniques including longitudinal studies with a large sample of individuals and several data gathering campaigns, different studies have demonstrated the existence of a pattern detailing an influence in the consumption of alcohol among the members of a social group, and specifically among adolescents [9,10].

Further studies have focused on particular issues related to these selection and influence mechanisms observed within networks of adolescents. For example, Light et al. [39] also found some evidence on both selection and influence mechanisms in terms of the onset to first alcohol use. The study described by Schaefer [40] focused the research on the selection mechanism, its dynamics and how it can lead to peer networks that can later be an enabling environment for influence. The influence mechanism has also been studied by comparing online and offline social networks [41] or the gender of the peers [42].

Other studies have considered the dynamics of friendship networks, alcohol use, and physical activity in conjunction [43], relationships combined with other individual characteristics like popularity [44], or the importance of the position of the adolescent in the friendship network [45]. The study described by Wang et al. [46] introduced the need to study the synergistic effects of both peer and parental influences for adolescent friendship choices and drinking behavior. In addition, some research has been dedicated to using simulated models in order to reproduce and study these network dynamics [47,48].

SNA studies have also been used for studying interventions. For example, the changes in the social ties of the individuals immersed in an addiction recovery treatment are studied in Kelly et al. [49]. It is clear, that SNA has demonstrated its value for conducting these kinds of studies and obtaining valuable information on the consumption of alcohol among adolescents in regard to their social ties.

### 2.3. The usual process for applying SNA to healthcare studies regarding alcohol consumption

Researchers usually obtain data from the object population of study by means of surveys drawn up using one of the various questionnaire authoring tools such as Google Docs [50], SurveyMonkey [51], Limesurvey [52] or Survio.com. These tools are very useful in creating surveys easily and obtaining the data collected in a structured format, but they cannot carry out a social network analysis, and so the data needs to be extracted and fed into other tools for this purpose. Moreover, during this process, the transformation and curation of the original data needs to be achieved. As this is a complex process, it usually carried out manually and is prone to errors as well as being time consuming.

Questionnaires used for health-related studies are usually put together from existing, standardized ones, or by adapting them to the specific purpose of the study, creating an ad-hoc questionnaire that has to be validated. In any case, questions in these questionnaires use to be grouped together in order to obtain a measure of the characteristic that the researcher wants to study. For example, the AUDIT (Alcohol Use Disorder Identification Test) questionnaire [53] assigns a "risk zone" depending on the answer to a group of ten different questions. In order to use the outcomes of this questionnaire, a calculation needs to be made to obtain the risk zone for each individual. This task is usually accomplished in a semi-automatic fashion, for example by inserting a formula in an Excel spreadsheet.

If a given researcher wants to include a study into the social relationships of the individuals, then a set of questions in the questionnaire must be dedicated to gathering this kind of information. These kinds of questions produce an output in the form of a matrix of $n \times n$ dimension (for $n$ individuals filling in the questionnaire) reflecting whether is there a social relationship (be it a colleague, friend, classmate, etc.) between each pair of individuals. These data are also difficult to manage by the expert, who will spend some time extracting them from the questionnaire output and introducing them into the social network analysis tool for later processing.

Social network analysis tools such as UCINET [54] use the data obtained by the expert from the questionnaire and introduced into the application in order to calculate different SNA metrics that may be relevant for the study at hand. The output from these tools is usually fed into other specialized visualization tools in order to gain an insight into the situation regarding the social structure.

The next section introduces how techniques from the semantic representation of knowledge in the computer can help to overcome the difficulties and shortcomings of the aforementioned usual approach.

## 3. Semantic technologies for representing social network analysis techniques

This section shows how semantic technologies may help to study the social relationships that may play a role as regards alcohol consumption in adolescents. First, a brief introduction to the field of semantic knowledge representation is provided and then an overview into previous efforts for adding semantic content to SNA is presented.

### 3.1. Semantic knowledge representation

The use of semantic knowledge representation techniques makes it possible for a computer to represent and process information at high abstraction levels using a formal structure representing the set of concepts and relationships between them for a given domain. This conceptual structure (called an ontology) can be processed by the computer and is the basis for the automated reasoning processes within computer-aided decision support systems that may help, for example, to assess alcohol use disorder risks or even design and implement intervention plans based on these data.

Semantic representation of knowledge has its roots in the field of Knowledge Engineering. In the mid-to-late 1990s the term "ontology" was coined and described as "a set of representational primitives with which to model a domain of knowledge or discourse. The representational primitives are typically classes (or sets), attributes (or properties), and relationships (or relationships between class members). The definitions of the representational primitives include information on their meaning and constraints on their logically consistent application" by Tom Gruber [55]. Knowledge models represented in ontologies have been built and exploited in many different domains and systems, but the burden of information that grows on the Web makes them focus on the problems associated with information management in this Internet service. This fact has made the set of standards, languages and tools used today known, in general, as "Semantic Web" [56].

A knowledge-based system built with semantic technologies may comprise several components:

- An ontology, which is the formal structure reflecting the concepts and relationships existing between them for the given domain of application. An ontology may be thought of as a dictionary plus a number of relationships between the concepts that permit their accurate, formal description or definition. An ontology stores the concepts needed for being able to talk and describe things within a given domain. The most common, standardized language for representing ontologies is the Web Ontology Language (OWL) [57].
- A knowledge base that uses the concepts from the ontology in order to represent a particular set of data. For example, the individuals participating in a social network and the relationships between them. The most usual formal structure to represent knowledge in a knowledge base is the Resource Description Framework (RDF) [58]. An RDF graph is a set of elements and links between them. The elements and links in the graph are individualizations of the concepts in the ontology (for example a concept in the ontology may be "person" and an individualization may be "student with ID:8775553") and the links reflect a particular situation or characteristic about a given element (for example the "name" for "student with ID:8775553" is "Peter Smith").

- A tool for querying and extracting information from RDF graphs, or even for creating new, different, graphs that focus on particular aspects or explicit new relationships. The usual language for doing this is SPARQL [59].
- If extra expressivity is needed, beyond what the OWL language can represent, a rule-based language can also be used to infer new sets of knowledge from a given RDF graph. The standard language for this is Semantic Web Rule Language (SWRL) [60].

### 3.2. Semantic representation of SNA concepts

Erétéo et al. [61] created an ontology called SemSNA and a knowledge-based system, Corese, that uses this ontology to ease social network analysis studies. This tool makes it possible to analyze the characteristics of heterogeneous online social networks. In this research, three main objectives were achieved: (i) to represent social networks in an ontology, (ii) to carry out a social network analysis using a knowledge-based system and finally, (iii) to detect and semantically label communities of online social networks.

This work introduced the notion of "semantic social network analysis" in an effort to use Semantic Web technologies for representing the social networks and the associated set of metrics and methods used for their study. This approach of "semantic SNA" allows the representation of complex relationships that include semantic content beyond what is common in "traditional" SNA studies, where relationships used to be considered as simple links (with no extra semantics) between nodes.

Other previous approaches to applying semantic knowledge representation to social network analysis [62,63], use a more conservative point of view, in which the traditional, simple link social networks is represented by means of Semantic Web frameworks such as RDF, using well-known ontologies for representing relationships, as is the case of FOAF (Friend of a Friend) [64] for representing friendship links or RELATIONSHIP (http://purl.org/vocab/relationship) for building more specialized relationships (for example familiar, friendship or professional relationships).

Whatever the approach, the use of semantic knowledge representation in SNA offers a plethora of possibilities for improving this kind of studies [65]. The following section shows the approach used in the present research work.

## 4. Building a semantic knowledge model for SNA analysis of adolescent alcohol use

As was stated previously, the objective of this research is to build a software tool that help to get valuable information on the alcohol consumption habits in adolescents in relation to their particular situation and social networks where they are immersed. Building such an application is a complex and time-consuming task. Several expert roles are involved in the process and all of them must collaborate very closely. The following expertise is necessary to accomplish this work:

- Knowing the dynamics of adolescent friendship networks, alcohol consumption habits and the way social connections may influence these habits.

- Knowing the techniques and tools from the Social Network Analysis domain and how they can be applied to networks in which nodes are people, as well as the concepts that relate SNA generic metrics to roles and characteristics of interpersonal relationships.
- Knowledge on the semantic representation of data to create the conceptual structures (ontologies) that will facilitate the management and exploitation of the information. In particular, the SNA concepts as well as adolescent personal data, friendships and alcohol consumption habits must be represented in the conceptual structure.
- Expertise in building software applications and visual user interfaces in which data are fed from the ontology and knowledge base and presented in an easy-to-understand way to the final user.

The psychosocial health professional is the central figure in this research. These professionals have the expertise about the problem for which the development is carried out and they will also be the final target users for the application. They also know about the role that interpersonal relationships among adolescents play in the modification of alcohol consumption habits and the main mechanisms and roles that are thought to be involved in the process. While it is true that there is no complete knowledge into how these mechanisms work, it is clear that the study of adolescent social networks is a good way of being aware of this information and may help to clarify these processes.

The joint work of psychosocial health care professionals and experts from the Social Network Analysis domain is needed in order to find the SNA measures that are of interest to evaluate, characterize and analyze the adolescent situation within his/her network of relationships.

In order to build an application able to automate the processing and analysis of the information, the role of knowledge engineer is also needed for building the necessary knowledge models (ontologies) that will store the conceptual structures reflecting the knowledge of the psychosocial health professionals and the SNA experts. Finally, software engineers are in charge of designing and building an application that makes use of these conceptual structures, performs the analyses and presents the results to the user through rich, interactive graphical user interfaces.

The work developed to achieve the previous set of development stages is summarized as a workflow in Fig. 1.

### 4.1. Initial steps and pre-meetings to obtain application requirements

In the initial phase of this research, a number of meetings were held among different experts with the four roles involved in the design and development of the application: psychosocial health professionals, social network analysis experts, knowledge engineers and software engineers. Previously to these meetings, the SNA experts searched for, and studied, literature about applying these techniques to substance consumption scenarios, and alcohol in particular. The knowledge engineers also studied basic notions of SNA and alcohol consumption habits, trying to find a number of existing knowledge models (ontologies) that had previously been constructed for similar domains.

In a first round of joint meetings, psychosocial health professionals introduced the problem of alcohol consumption and the need for tools able to monitor and analyze the situation of alcohol consumption in a school environment. The social network analysis experts introduced and showed how SNA could help to find and study different useful measures that describe the relationships that an individual holds with other peers in his friendship networks. They gave examples of how SNA could be used in the case of alcohol consumption according to the related literature they had found. During the meetings there was continuous feedback between the psychosocial health professionals and the SNA experts, clarifying concepts and possible application scenarios.

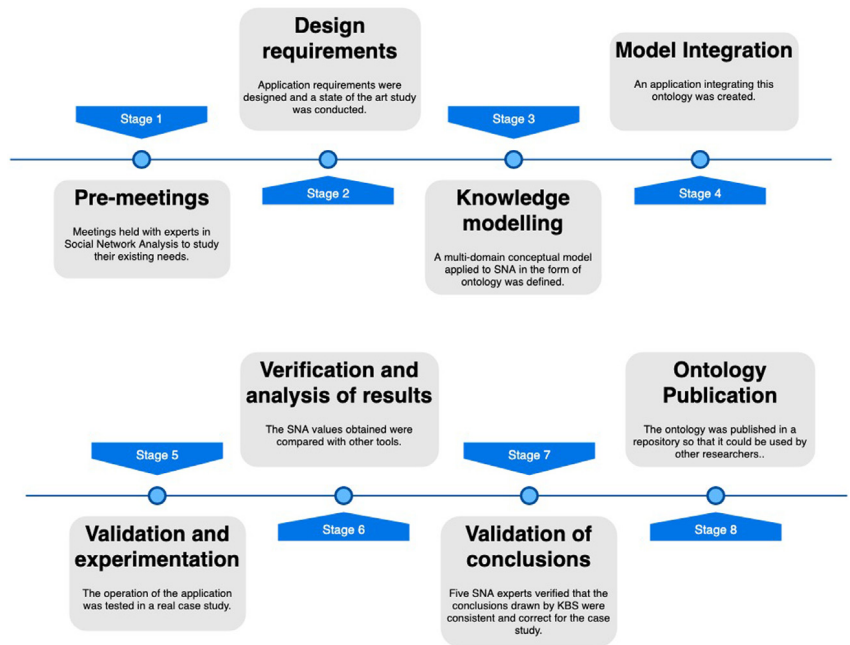

**Fig. 1.** Project workflow diagram.

During this first meetings, the knowledge engineers asked questions to both groups of experts, promoting discussion about the different concepts that appeared, with the aim of obtaining clear and concise descriptions and definitions needed to build a formal conceptual model (ontology).

After this first meeting, the different groups worked separately, communicating occasionally with the other groups if they needed some further explanation about a given issue. In a second round of joint meetings, the psychosocial health professionals presented the kind of functionalities that they would like to ask the application to provide, by stating a series of specific needs that were captured as questions:

– What is the alcohol consumption level in a classroom?
– How dense is the friendship network? How many groups can be found in the class when considering these friendship links?
– Do these groups contain students with similar alcohol consumption levels?
– What is the risk of alcohol use disorder for each student?
– Which student/s or group can be an influence for a given one?
– Given an individual, which ones can be influenced by him/her?
– Does the alcohol consumption level relate to family wealth?
– Is the student with a higher alcohol consumption level the most popular (that is, the one having a strategic position to contact other students) from the friendship standpoint?
– In general, which profile do the most popular students in the network have?
– Is alcohol consumption a good measure of popularity among students?
– Are there mediator figures in the network? (People who act as connectors between more or less separated groups) If so, what relationships are there between popular student/s and mediator/s?

– Does it make a difference if the boys' network is studied separately from the girls' network?
– Who is the central student/s when considering alcohol consumption? Is it the same one/s than when considering the friendships? And, in general, how do the friendships and consumption networks compare?
– Do people choose the same people they name as their best friends as those for going out for a drink with?
– How the self-efficacy perception of the student relates to his/her position in the network, level of alcohol consumption, etc.?
– How the quality of life perception relates to alcohol consumption or number of friends?

By discussing these issues, the knowledge engineers refined and completed their conceptual models and the software engineers built the requirements for the application and designed the graphical user interfaces and the nature of the interactions to be implemented. Then, a more traditional software development phase started, following the usual stages used for programming web applications.

### 4.2. Design requirements for the application

As a result of the previous set of meetings, the application functionalities were stated, as represented in the diagram in Fig. 2. First, the application loads the set of data coming from the responses to a questionnaire used for gathering information. The appropriate social networks are generated and a number of analyses are performed that give, as a result, a series of reports and a number of graphical representations that will be conveniently shown to the user.

The application shows basic data and a report-like description of any individual in the network regarding his/her alcohol consumption status, especially concerning the risk of alcohol use disorder. In addition, the application shows, for each individual, who can act as an influencer for him/her, in order to evaluate the risk of being influenced for acquiring higher alcohol consumption levels.

The application also allows to show and study the level of alcohol consumption in relation to genre and environment, to show and report the relationship that may hold between alcohol consumption, socioeconomic status, self-perception and self-efficacy or polydrug consumption.

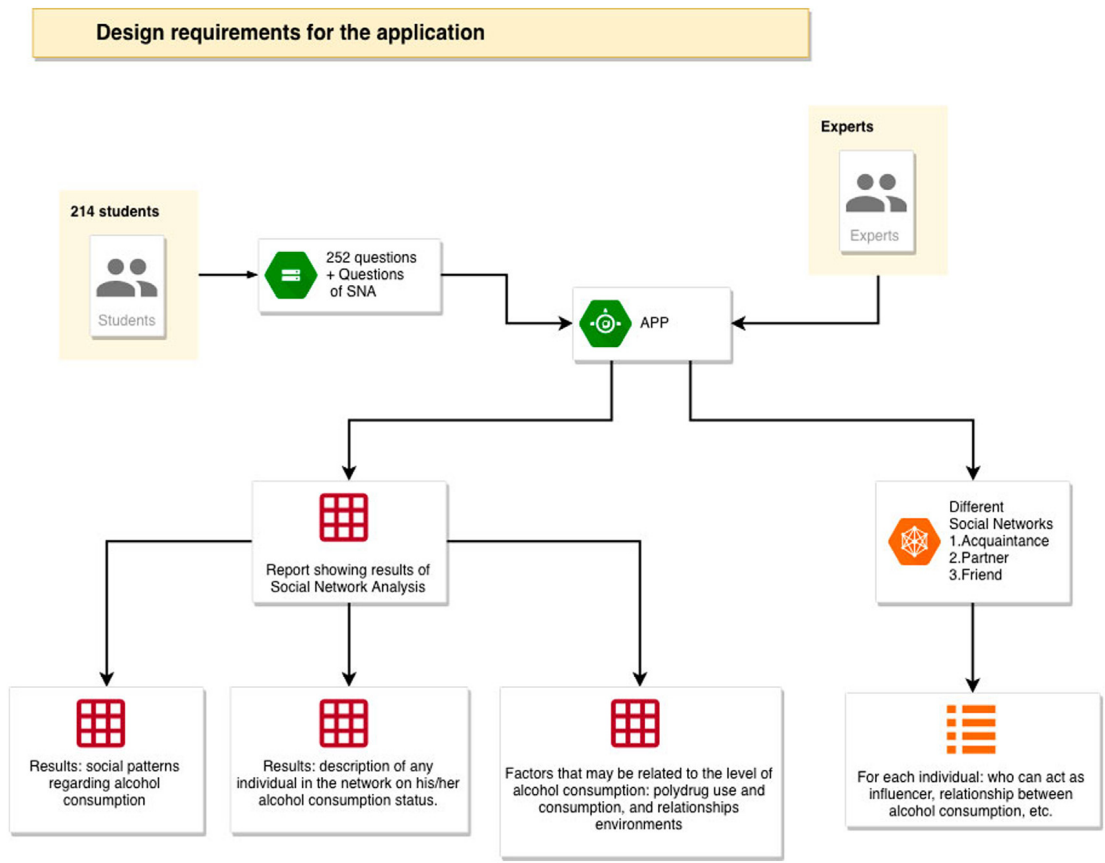

**Fig. 2.** Design of the requirements for the application.

The application must show the information in such a way that any health professional or researcher can understand it, without the need to know anything about SNA techniques or terminology. This is possible because of the use of the ontology and the processing structures used, that isolate the user from the underlying technical SNA measures and methods. The interaction with the tool and the information presented consists on commonly-used expressions when describing adolescent characteristics, friendship, family relationships and alcohol consumption.

### 4.3. Data gathering with a questionnaire

In order to help the development process and test the validity of the approach, a questionnaire was put together for collecting data to be used in the application in a real scenario. The questionnaire was drawn up using some standardized, validated tests and a number of adapted questions. The whole questionnaire can be consulted in https://doi.org/10.17632/ykf85w9h83.1. The questionnaire was designed by the psychosocial health professionals, with some help from the SNA experts for the network generating questions. Two main parts can be distinguished in it: one of them is dedicated to obtaining data on personal, individual and family situation and the other one is aimed at obtaining information on the social links between individuals. Three environments were considered for studying the social networks: the friendship and peer network (the class and school mates), the external network (friends outside the school) and the family network.

Some questions were devoted to obtain the basic data of the student: full name, date and place of birth, and gender. In order to obtain the information to characterize the individual, his socioeconomic profile and alcohol (and other drugs) consumption habits, a number of standardized tests were used or adapted, as shown in Table 1. The AUDIT test with 10 questions was chosen due to its suitability for the purposes of the application: it allows to characterize the individual's alcohol consumption risk by using only ten items, and so it would not overburden the full questionnaire. The FAS II affluence scale was chosen due to its widespread use and its suitability for getting a valid and reliable measure of socioeconomic status in developed countries. The ESTUDES items were chosen because it is tailored to the gathering of data about substance consumption among Spanish adolescents. Finally, the KIDSCREEN 27 version was used because it allows to obtain a measure of the quality of life perception in adolescents using a minimum set of items.

Social relationships were captured by a number of questions about friendship and alcohol consumption. The *friendship network* was built up using a question with a Likert-like scale of five values for quantifying the intensity of the link to any other student in the same class (see Section 5 for details); three levels of relationship are calculated based on the response to this question: acquaintance (weakest), partner (medium) and friend (strongest). Another question is dedicated to obtaining the links to other students as regards alcohol consumption habits by asking if the respondent would go out for an (alcoholic) drink with the other class mates, thus building up the *consumption network*. These two questions help to know, for example, if the individual is willing to drink with the same people they consider as their best friends or not, which might imply a number of consequences from the point of view of influence processes.

A question was dedicated to identifying friendship relationships outside the class/school. Data from this question is an indicator of how many friends the adolescent has outside the educational environment, what could also be relevant for his/her alcohol consumption behavior. This *external network* cannot be used for the SNA analysis because individuals do not belong to a common social environment.

**Table 1**
List of standardized test used to do a social network analysis.

| Standardized test | Description and utility |
| --- | --- |
| **AUDIT** (Alcohol Use Disorders Identification Test) [53]. | This test is used to detect problematic levels of alcohol consumption, or dependence. |
| **FAS II** (Family Affluence Scale II) [66]. | Used for assessing family wealth. |
| **ESTUDES** (Poll into drug use in secondary schools in Spain) [5]. | A biannual study for gaining an insight into behaviors and attitudes regarding substance use. It is a test that includes questions about the consumption of different substances. Those related to alcohol were discarded as they were obtained in other parts of the main questionnaire. |
| **KIDSCREEN-27** (Health Related Quality of Life Questionnaire for Children and Young People and their Parents) [67]. | Used for establishing the quality of life perception of the adolescent based on five scales: physical and psychological well-being, autonomy and parental relationships, peers and social support and school environment. |
| **Self-efficacy** [9] (Spanish adapted version). | Assesses the belief of the adolescent as to his/her own capacities to achieve different goals, especially facing stressful situations. |

Finally, some questions were dedicated to alcohol consumption habits within the family network. The aim of this part of the questionnaire was to characterize the alcohol consumption habits in the family, trying to study whether there is any relationship to the adolescent's own habits.

### 4.4. Knowledge modeling: Semantic knowledge representation of the domain

The ontology to be generated had to represent concepts from three domains in an integrated way:

- To represent knowledge in the field of people.
- To represent knowledge in the field of questionnaires.
- And to represent knowledge in the field of Social Network Analysis.

Before carrying out the modeling of knowledge in the form of ontology and after studying the art of the technologies applied in the field of SNA, it was decided to reuse existing ontologies and add the relevant classes and properties to achieve the initial objective of this research.

The ontology used to represent the concepts related to people and their characteristics was FOAF [68]. In the domain of questionnaires, an ontology found in the work of Alipour-Aghdam was used [69]. And finally, the basic ontology used to conceptualize the concepts of SNA was SemSNA, by Ereteo [70]. These separate ontologies did not have all the concepts necessary to achieve the objective of the system. This is why a new ontology was constructed, taking as a basis the previously mentioned ones, thus filling the existing shortcomings in them.

A view of the classes in the ontology developed (called *OntoSNAQA — Ontology, Social Network Analysis, Questions & Answers*) can be seen in Fig. 3. It has three different domains: People, Questionnaire and Social Network Analysis. The ontology is available in a Mendeley[1] public repository. This ontology has the following metrics:

- 661 axioms
- 409 logical axioms

- 210 declaration axioms
- 74 classes
- 133 object properties

  - 25 properties to represent the different relationships between the Person and the Questionnaires and the Person and the Social Network Analysis.
  - 42 properties allow Questionnaire concepts to be related to Person concepts and SNA Concepts.
  - 66 properties needed to represent all relationships between the social network analysis with the domain of People and the domain of Questionnaires.

- 24 data properties

The ontology was built with the Protégé ontology editing tool [71]. The main classes observed in Fig. 3 are described as follows:

- Person: It is a reused concept from the FOAF ontology, to which it has been necessary to add new data properties and two new sub-concepts. Below are the subclasses that have been necessary to generate in the new ontology:

  - ClassSchoolData: This concept establishes information on a class at a primary school because, in our case of study, this information about the people is useful to create different relationships automatically. Other subconcepts like *AcademicCategory, Course, CourseLevel, GroupOfClass*, and *School* have been established in this concept.
  - ClassOnSchool: This concept allows a complete class at a primary school to be represented. Instances of this class can relate to different concepts about ClassSchoolData. There is an example of an instance of this class in Fig. 4.

- QuestionnaireDefault: These concepts were based on the ontology of Alipour-Aghdam [69], but it was also necessary to add new sub-concepts, namely three were added:

  - QuestionSNA: It is a sub-concept of *Question* class that helps to researchers to represent questions related directly to the networks. For example, a question of this type could be: *How much time do you spend with any people of the list shown below?* This question will be repeated with all the members in a network. If the questionnaire is to be completed by 30 students in a class, this question will be asked 30 times to each student.

---

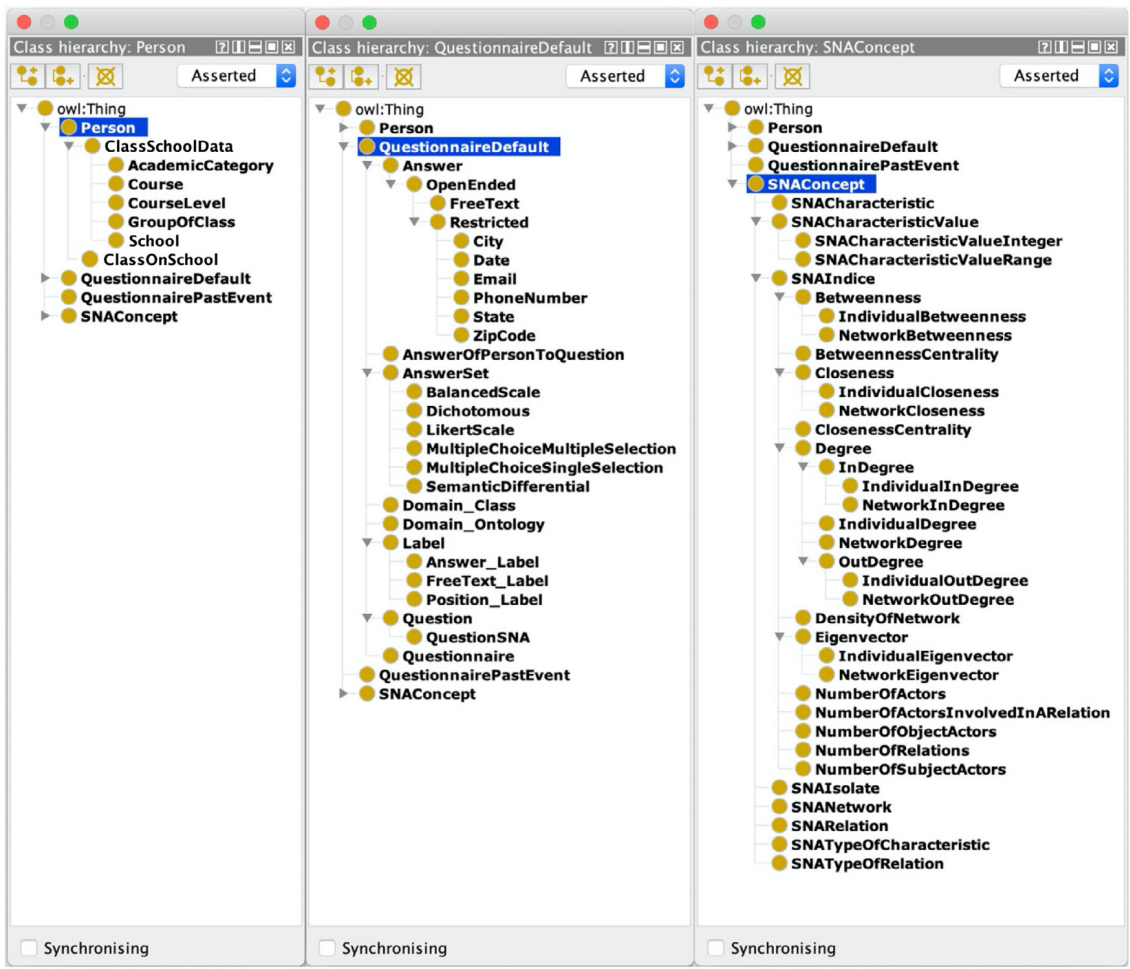

**Fig. 3.** Classes of the ontology OntoSNAQA.

– QuestionnairePastEvent: This concept is necessary to represent a Questionnaire completed at a certain time. The questionnaires could be completed by users as many times as the researcher needs throughout his or her research. For example, a user can complete a questionnaire entitled Breathalyzer risk questionnaire in 2016 and he or she can do it again in 2017. Thanks to this, it is possible to relate to the different members of a network with the two events of the same questionnaire at different times.

– AnswerOfPersonToQuestion: This concept represents an answer of a user to a certain question in a certain questionnaire completed at a certain time. This concept allows the concepts of: Person, QuestionnairePastEvent, Question and Answer to be related.

There is an example of different instances that interact with each other in the ontology developed in Fig. 5.

• SNAConcept: The conceptual model by Ereteo [61] had a series of classes through which typical concepts could be represented in an analysis of a social network like SNAIndice, Degree, Betweenness, Closeness, etc. However, the needs of the model presented required a larger number of concepts to be added to make it more complete and meet the requirements of the initially proposed objectives. This is why it has been necessary to add 24 new classes to the initial model to achieve to conceptualize more measures of SNA like Betweenness, Centrality, Closeness of each person and each network, and others.

In addition, the ontology has seven rules in SWRL language and eleven queries in SPARQL Update. Rules in SWRL allow the KBS to do the next tasks:

• Linking a person with a network and a questionnaire at a certain date.
• Creating relationship instances between different people in a network.
• Assigning a characteristic to a person in a network in relation to his/her answers to a questionnaire.
• Creating instances of the class SNACharacteristic.
• Finding which answer of a person is related to a SNACharacteristic.
• Creating instances of SNAConcepts of a person in a network.
• Creating instances of SNAConcepts of a network in a network.

Moreover, SPARQL Update queries help the KBS to calculate different values of Social Network Analysis metrics and to assign this value as a data property of a SNAConcept in the ontology.

The use of ontologies and RDF graphs to represent the information greatly facilitates the handling and integration of data. The results from the questionnaire passed to the students are stored as RDF graphs with rich semantic content. Basic individual data such as the name, date, and place of birth, gender, etc., are stored as simple relationships from individuals to simple datatypes (as string, float, date time, etc.). Responses to the different tests are also stored as numerical values (for example, the AUDIT score, the FAS II affluence level test, etc.). A separate knowledge structure relates these numerical scores to the qualitative value that corresponds to each value interval. As a result, information can be presented to the user in a more familiar way without any need to know how these tests results are graded.

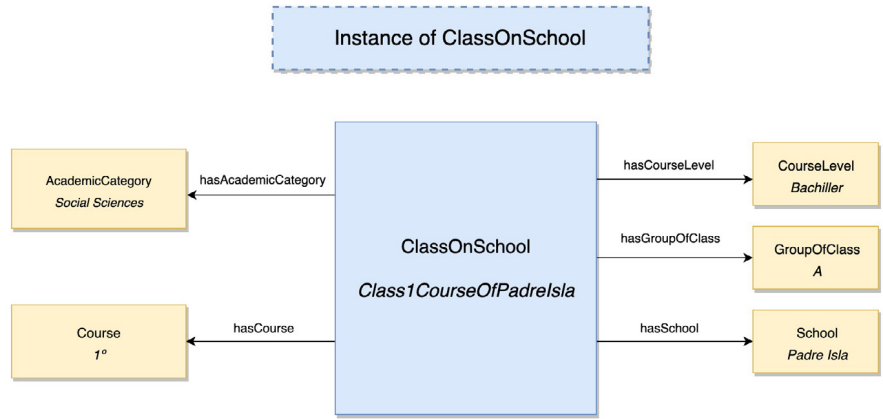

**Fig. 4.** Example of ClassOnSchool instance.

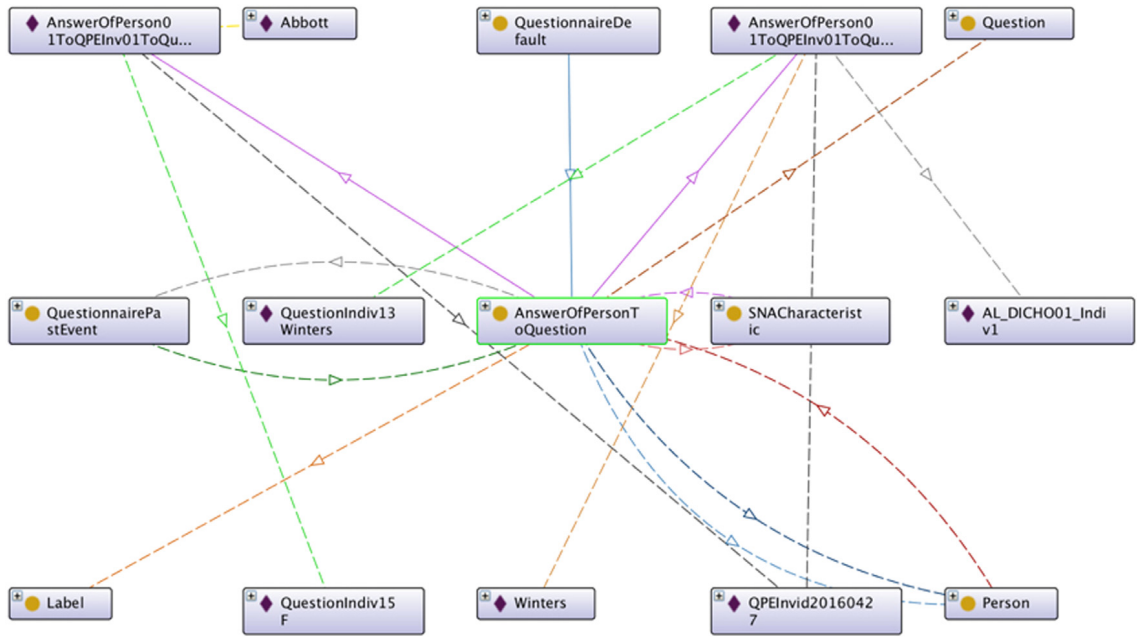

**Fig. 5.** Conceptual representation on a person who has answered a questionnaire at a certain time using OntoSNAQA.

Two main social networks are represented initially from the data of the questionnaire. The first one is the friendship network, in which there is a relationship between every two students based on the response to the corresponding question. This relationship is directed (a person says he/she is a friend of another one) and weighted (the person declares the strength of this friendship link). The second one is the consumption network, compiled with the results from the question in which students are asked to tell if they would go out for a drink with each of their classmates, in this case the relationship is directed but not weighted. From these two main social networks, represented as RDF graphs, a number of others are calculated using the SPARQL language. For example, only the friendship relationships graded with the highest scores for the strength of the link may be used to compile a network of "friends", the direction or the reciprocity of the relationship can also be taken into account, etc.

Then, the metrics and algorithms from SNA are implemented in all the networks that have been created. The results from these metrics are stored in the knowledge base using the concepts in the SNA ontology, that is, the corresponding graph containing the social network is annotated with a series of properties related to SNA measures, following the idea of named graphs [72].

Like for social networks, a number of properties are also calculated and associated with individuals using the results of the queries in the questionnaire and rules expressed in SWRL. Furthermore, the roles that each individual plays in the different social networks are associated with his/her profile description based on their characteristics and the SNA metrics calculated for each of them in the context of each of the different networks in which he or she is involved. It is possible to appreciate the difference between Social Network Analysis made using our app based on semantic technologies and others in Fig. 6. As can be seen, not using knowledge models for automating data management leads to a lot of manual handling (dotted, red lines in the figure) that may lead to errors.

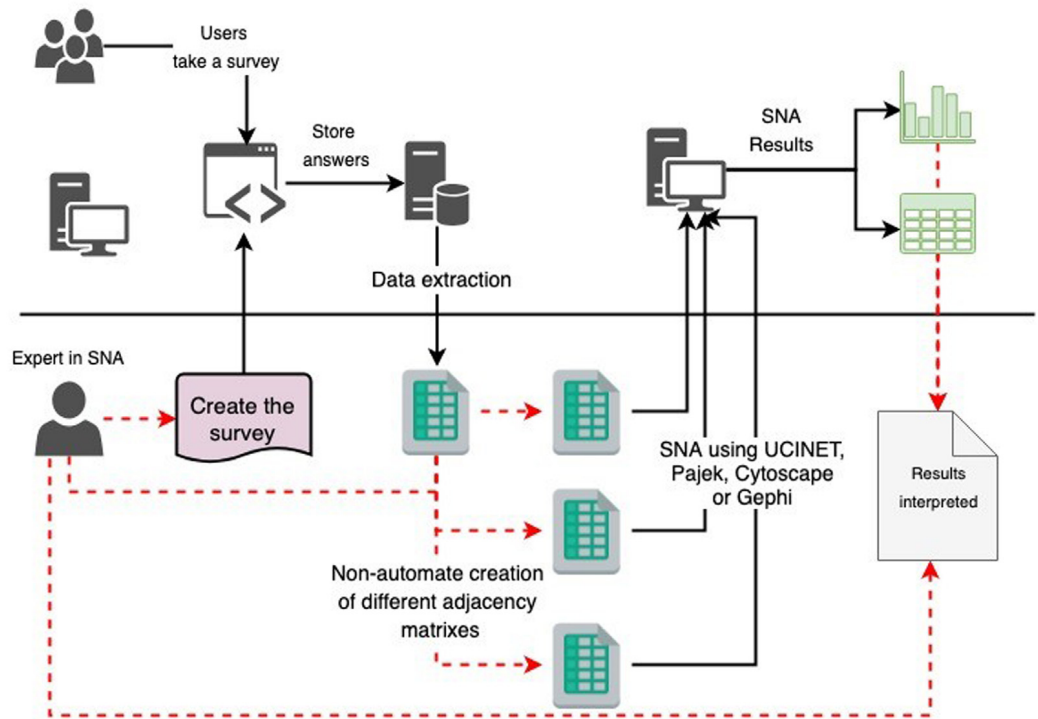

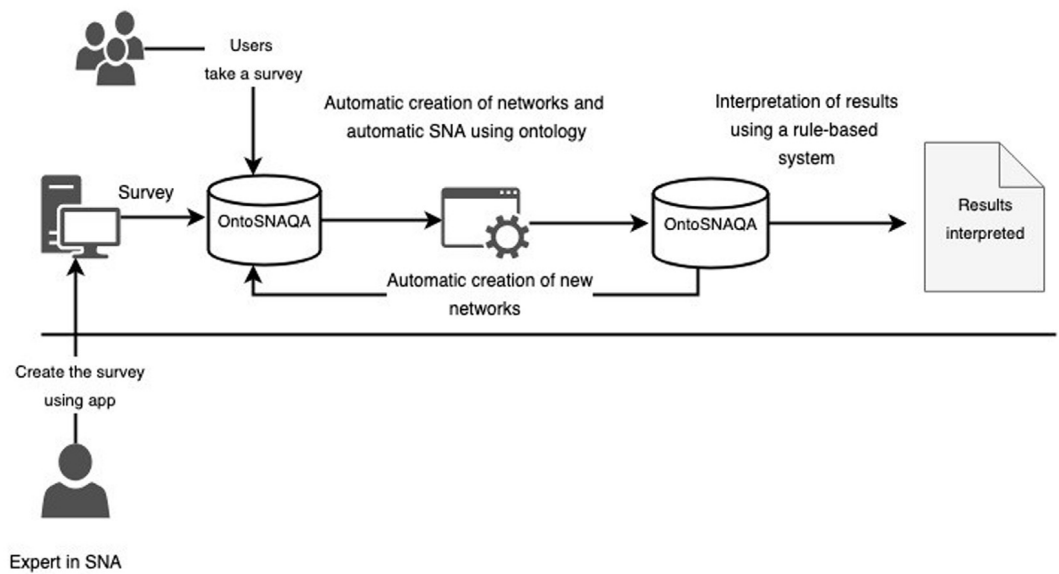

**Fig. 6.** Differences between performing a social network analysis applying semantic technologies and without using them.

*4.5. Model integration: Creating an application that makes use of ontology*

The last step of the development stage, prior to validation, was to create an application that would make use of the ontology generated, OntoSNAQA. To do this, a web application was programmed using HTML5, CSS, PHP, Javascript and various Web services that allowed the ontology to be used more easily.

The application allows the user to create a questionnaire with the necessary tests and items. Within the questions, there is the possibility of indicating which one corresponds to a question that generates a social network. In this way, the ontology will relate instances of type Question with instances of type SNAConcept.

Once the questionnaire is completely generated, the expert has the possibility of importing the users who will fill the form.

**Table 2**
Possible statements and their corresponding weights for the social network generating question.

| Description/degree of contact | Weight |
|---|---|
| *We never spend time together* | 1 |
| *We sometimes spend time together* | 2 |
| *We use to spend quite a lot of time together* | 3 |
| *We are almost always together* | 4 |
| *We are always together* | 5 |

**Table 3**
AUDIT risk level scoring.

| Risk level | Intervention | AUDIT test score |
|---|---|---|
| Zone I | Alcohol education. | 0–7 |
| Zone II | Simple advice. | 8–15 |
| Zone III | Simple advice plus brief counseling and continued monitoring. | 16–19 |
| Zone IV | Referral to a specialist for diagnostic evaluation and treatment. | 20–40 |

After this step, the population to whom the research is directed must respond to the questions so that, finally, the expert can execute the analysis step. In this step, the application is in charge of applying all the SWRL rules and SPARQL queries that have been explained previously in this document in order to obtain a social network analysis and its conclusions.

## 5. Experiments and results

With the main objective of testing whether the conceptual model in the OntoSNAQA ontology is able to obtain correct conclusions on social network analysis and assessing the validity of the approach for the healthcare professionals, a web-based application that uses this ontology was compiled.

In this application, once all the information has been processed and stored in the OntoSNAQA ontology, the system applies all SWRL rules automatically using the Pellet reasoning engine [73], and the SPARQL Update queries. Then, the user can visualize and interact with the results obtained using the graphical user interface (GUI) of the application. The amount of data that can be accessed is big and complex, and so the semantic constructs are used again to obtain near-natural language descriptions of the particularities of the individual.

To test the application, a study was carried out in a real educational environment. A sample of 195 adolescents aged between 16 and 19 years old, who were in their first or second year of post-compulsory secondary education, were selected at four public high schools in the region of El Bierzo (León, Spain), in 2017. A total of 7 classrooms were involved. Parental consent was obtained for all participants, who were recruited through convenience sampling. The total sample consisted of 53.85% females ($n$: 105) and 46.15% males ($n$: 90), who presented a mean age of 17 years old ($SD$: 0.82). Students in their first year of post-compulsory secondary education accounted for 53.85% of participants, while those in their second year for 46.15%. The socioeconomic data indicated that 39.29% of participants presented a medium–low level, whereas 60.71% presented a high level.

The class with the higher number of students was chosen to be studied with the aid of the software applications. In this classroom, there were a total of 38 students from whom different sociodemographic data were collected. Three networks were obtained: friendship, partnership, and acquaintances. In order to obtain these networks, adolescents had to reply to this generating question:

*"How much time do you spend with each of the following class-mates?"*

Adolescents were able to choose the degree of contact with each of their peers by choosing one out of five possible statements. Each of these statements has an associated weight (see Table 2) that is later used to qualify the relationship.

When both individuals used a statement having a weight of 4 or 5 to define their level of contact they were considered "friends", that is the strongest link two adolescents can have. This definition is stored in the ontology and so helps to disambiguate and precisely define what the concept of "friend" entails in this scenario.

Fig. 7 shows the information displayed for the main social network of friends. In this image, circles represent individuals. Pink (light gray) colored-ones are girls while blue (dark gray) ones are boys. The size of the circle represents, in this view, the AUDIT risk zone as established from the questionnaire response for each student (bigger circles mean higher levels of alcohol consumption), see Table 3 for audit zones scoring.

The user can click on any of the circles and the detailed information for that student will be shown in the left-hand part of the window (see Fig. 8). The data than can be obtained includes:

- Anonymized name, age, place of birth, gender and school.
- Family affluence, as obtained in the FAS II questionnaire. Both the qualitative value and the quantitative test result are shown.
- Friends outside classroom, friends that each user has but that are not within the school environment.
- Drinking mates outside classroom, friends with whom they would drink alcohol but who are not within the school environment.
- Family drinking frequency, how often they drink alcohol with their family members.
- AUDIT test results, both the risk zone and the test result are shown.
- KIDSCREEN test result.

One of the interesting things that can be noted at first sight in this view is, for example, the fact that the most popular individual (the one that is named as friend by more people) is also the one with a greater AUDIT score (the individual anonymized as "Gay"). With the aid of different menus and items the user can easily navigate to any other graph generated during the data processing phase.

Fig. 9 represents the window showing more detailed information for a given individual. This window shows the basic personal data for the individual and a visual representation of the calculated alcohol use disorder risk level. The individual responses to the different tests filled by this adolescent can be obtained by clicking on the "Questionnaire responses" link.

Two natural language descriptions are also generated from the knowledge base and presented to the user. The first one describes the particularities of the individual from the point of view of the friendship network in which he or she is involved, while the second indicates the alcohol consumption habits. The questionnaire data and the concepts defined in the ontology are used in order to calculate the different values and build these sentences.

Some social measures such as the popularity, capacity of influence or mediation are also presented based on the positions and connections the student has in different social networks. The definition of the different levels of popularity, role as mediator and level of influence are defined in the ontology by establishing ranges and formulas involving different SNA measures like *indegree*, *betweenness* or *closeness*, and also the weights obtained for the social network ties in Table 2).

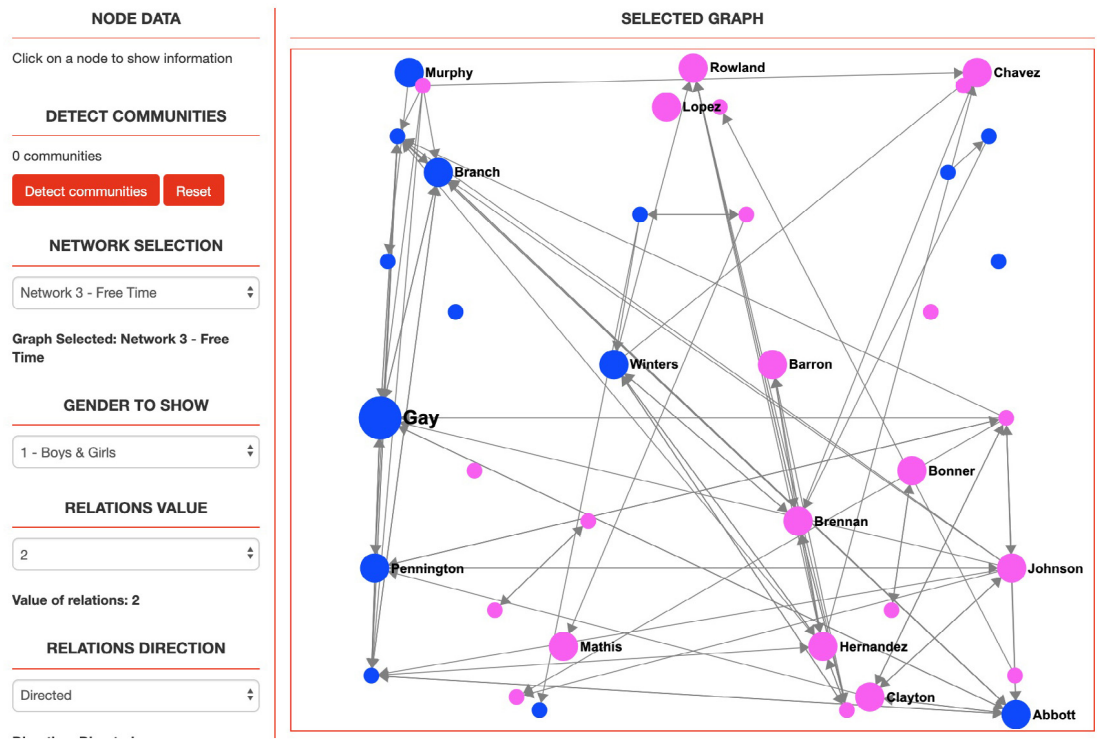

**Fig. 7.** Application screenshot showing the social network of friendship relationships.

Finally, the user can find a number of mediators that can facilitate the information flow towards the current student, or some individuals that may influence him. This information is calculated by studying the social connections of the individuals and the characteristics of these connected friends. This individual can also be shown in any of the different social graphs in which she appears by choosing the graph in the drop-down box at the bottom.

The knowledge base that resulted from the application of the knowledge structures in the ontology has the following metrics:

- 8874 instances
- 19,621 ClassAssertion
- 66,253 ObjectPropertyAssertion
- 964 DataPropertyAssertion

### 5.1. A use case for the system

The application can be used as a platform for performing new alcohol consumption studies in school settings. The steps to follow by an interested user would be:

- Create a profile for the study.
- Provide a list of the names of the students involved in the study in order to populate the network generating questions in the questionnaire.
- Get the questionnaire filled by the students by using the online tool.
- Ask the system to perform the data curation and analysis.

- Use the graphical user interface in order to visualize and navigate the different windows where the information generated by the application is shown.

The application can be installed from scratch as a standalone web server devoted to the given study or may be used as a general platform where different studies can be achieved by creating different profiles in the same webserver (this latter use case is the one that has been described previously).

The questionnaire to be used would be the same as the one developed for the experiments and previously described. It is a fully automated online questionnaire and the user has not to worry about any kind of manipulation of the data, as all information is automatically passed to the analysis module.

### 5.2. Validation of the system

A two-step validation of the system was achieved: first, the results of the different formulae and methods used in the definitions of the SNA measures in the ontology were compared to the results of the more popular SNA software suits. Then, a validation of the different analyses performed by the system was achieved.

Comparing the values obtained with those obtained for the same set of data using the most common SNA tools: UCINET [74], Cytoscape [75], Pajek [76] and Gephi [77] (and a large amount of manual processing), the results obtained for the different SNA measures and values were exactly the same.

In a second step, a validation of the results from the analyses drawn by the system was achieved by validating each of the possible reasoning processes performed by the application according to the information in the ontology. Each of the possible combinations of input data needed to test all the possible analysis processes and definitions in the ontology was generated in an artificially constructed friendship network, and the results obtained by the automated analysis of the different cases were validated by the SNA experts and the healthcare professionals.

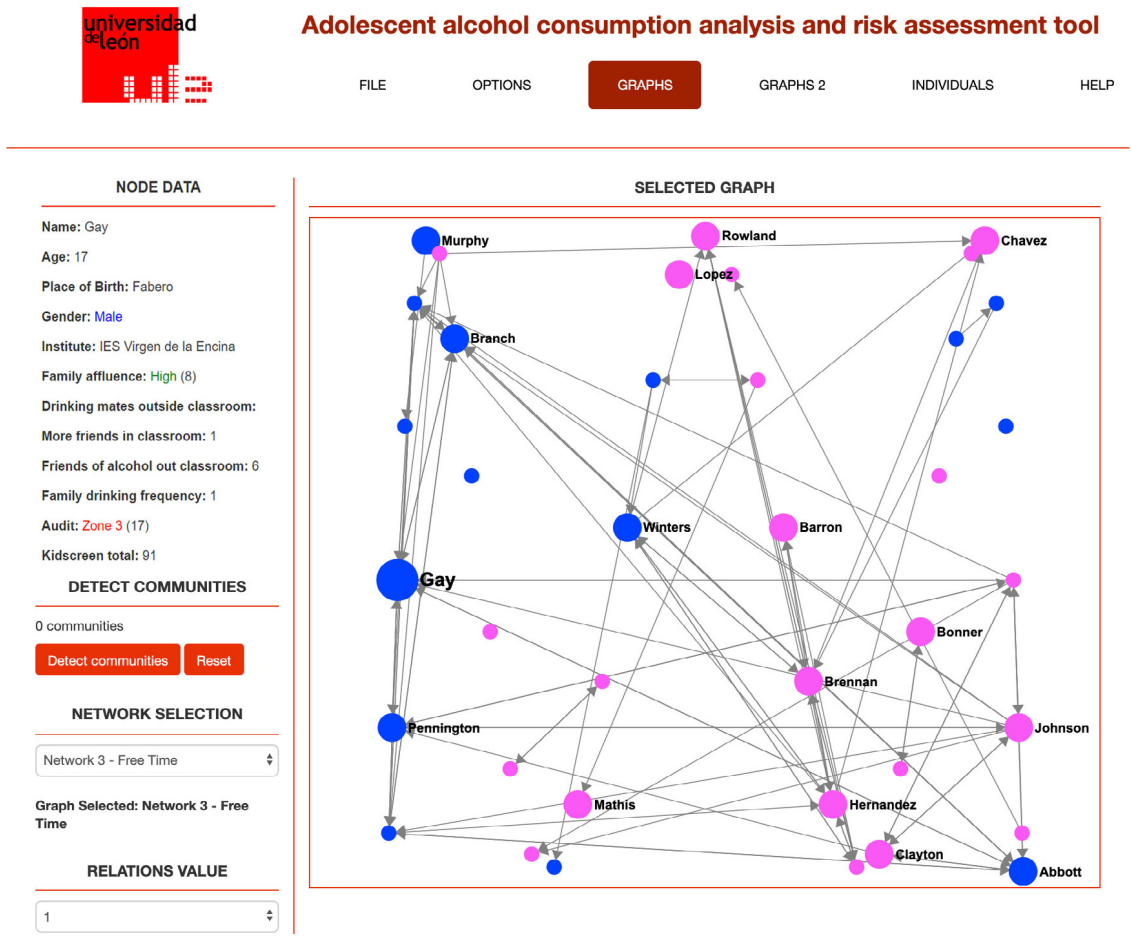

**Fig. 8.** Application screenshot showing individual detailed data.

## 6. Discussion and conclusions

The development of a fully automated tool for studying alcohol consumption habits in the context of the social networks where the individual is immersed is a complex work, as it demands the construction of complex knowledge models able to capture the expertise from different professionals. This interdisciplinary nature of the research demands a great effort from all the involved persons, but the outcome is worth it. The final system will serve as a complete, automated, application for gathering, analyzing and presenting a complex set of data about the alcohol consumption habits of adolescents and the social environment where all it takes places, obtaining valuable information about the current situation and some important issues that can be used for obtaining advice about risky situations or possible opportunities for interventions, that could even be based on and use the information of the same social networks where the adolescent is immersed [78].

The advantages of the developed tool are clear. The management of data is fully automated, which avoids human errors due to the manual handling of this information. The time spent on the studies, from data gathering to analysis and visualization of the results is drastically reduced, allowing the user to focus on what is relevant in the study (the concepts regarding the alcohol consumption situation of the different individuals) and not in the processes and formulae to be used to obtain the results. Using this tool, any health care professional or even a secondary school manager with knowledge of alcohol consumption problems can use the system and obtain relevant, meaningful results without the need of knowing about the techniques used to get them. In this sense, the tool can be used for gaining insight into the status of a classroom or school regarding alcohol consumption and obtaining warnings on the situation of given individuals that can be at risk of developing harmful consumption habits.

A number of improvements or modifications may be implemented in the application as future work. One of them is the possibility of including capacities for managing and analyzing the evolution of friendships and alcohol consumption over time, that is, being able to process questionnaires that are delivered periodically to the same group of students.

The tool can also be extended for developing intelligent decision support systems to be used for the planning and implementation of interventions aimed at individuals or groups as regards their alcohol use behavior.

The ideas and tools presented in this paper can be applied to other similar scenarios beyond the classroom and school environments. They could be used in adult populations and for much bigger communities, for example in the context of an online drinking cessation intervention. However, it is worth to mention that the algorithms and analyses performed for the current, small-size scenario (the classroom) would not scale well when dealing with communities of hundreds or even thousands

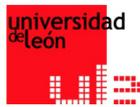

## Adolescent alcohol consumption analysis and risk assessment tool

FILE     OPTIONS     GRAPHS     GRAPHS 2     INDIVIDUALS     HELP

### INDIVIDUAL SELECTION

Abbott
Ayers
Barnett
Barron
Bonner
Branch
Brennan
Chavez
Clayton
Crosby
Dorsey
Ewing
Foster
Gay
Glenn
Hall
Hernandez
Johnson

### INDIVIDUAL SELECTED

**Name:** Barron

**Age:** 19

**Gender:** Female

Other filiation data

Questionnaire responses

**Overall alcohol abuse risk:**

**Personal and friendship highlights**

Barron is a 19 year old. She is not a popular girl among her friends who are girls in their majority.

Barron declares to have 1 friends and she is considered friend by 4 persons.

**Alcohol consumption highlights**

Barron has a medium level of alcohol consumption. Her most close friends have similar levels of alcohol consumption but she may be influenced by people with high levels.

Alcohol consumption habits: She was 14 when she tried an alcoholic drink for the fist time. She declares to have alcoholic drinks 2 to 4 times a month, with five or six drinks per occasion. The places where she goes for a drink more frequently are pub/disco.

AUDIT Score: 10 (Zone 2)

**Social metrics:**

Popularity: Low

Role as mediator: Low

Level of influence: Low

**Tools:**

Find mediators for Barron

Find influencers for Barron

**Show Barron in graph:**

Select a graph

**Fig. 9.** Application showing an individual.

of users. In these cases, emerging techniques like node or community embedding [79] should be applied in order to create more compact representations of the characteristics of nodes and communities that would ease the implementation and execution of analysis algorithms, solving the computational infeasibility issues that would arise.

### Declaration of competing interest

The authors declare that they have no known competing financial interests or personal relationships that could have appeared to influence the work reported in this paper.

### Acknowledgment

The work presented in this paper was supported by Junta de Castilla y León [grant number LE014G18].

### References

[1] M. Syed, I. Seiffge-Krenke, Personality development from adolescence to emerging adulthood: Linking trajectories of ego development to the family context and identity formation, J. Pers. Soc. Psychol. (2013) http://dx.doi.org/10.1037/a0030070.

[2] S.-J. Blakemore, K.L. Mills, Is adolescence a sensitive period for sociocultural processing? Annu. Rev. Psychol. 65 (2014) 187–207, http://dx.doi.org/10.1146/annurev-psych-010213-115202.

[3] R.M. Viner, E.M. Ozer, S. Denny, M. Marmot, M. Resnick, A. Fatusi, C. Currie, Adolescence and the social determinants of health, Lancet. 379 (2012) 1641–1652, http://dx.doi.org/10.1016/S0140-6736(12)60149-4.

[4] WHO, Global Status Report on Alcohol and Health 2014, WHO, 2016, http://www.who.int/substance_abuse/publications/global_alcohol_report/en/. (Accessed 18 June 2018).

[5] Estudes (2014): encuesta sobre el uso de drogas en enseñanzas secundarias. Ministerio de Sanidad y Consumo., 2014, http://www.pnsd.mscbs.gob.es/profesionales/sistemasInformacion/sistemaInformacion/encuestas_ESTUDES.htm. (Accessed 19 June 2018).

[6] J. Pons, S. Buelga, Factores Asociados al Consumo Juvenil de Alcohol: Una Revisión desde una Perspectiva Psicosocial y Ecológica, Psychosoc. Interv. 20 (2011) 75–94, http://dx.doi.org/10.5093/IN2011V20N1A7.

[7] I. Sánchez-Queija, C. Moreno, F. Rivera, P. Ramos, Tendencias en el consumo de alcohol en los adolescentes escolarizados españoles a lo largo de la primera década del siglo xxi, Gac. Sanit. 29 (2015) 184–189, http://dx.doi.org/10.1016/J.GACETA.2015.01.004.

[8] M. Steketee, H. Jonkman, H. Berten, N. Vettenburg, Alcohol use among adolescents in Europe, Enviromental Res. Prev. Action (2013) 351.

[9] T.A.B. Snijders, C.E.G. Steglich, M. Schweinberger, Modeling the co-evolution of networks and behavior, Longitud. Model. Behav. Relat. Sci. (2007) 41–71, http://en.scientificcommons.org/54922763.

[10] D.W. Osgood, D.T. Ragan, L. Wallace, S.D. Gest, M.E. Feinberg, J. Moody, Peers and the emergence of alcohol use: Influence and selection processes in adolescent friendship networks, J. Res. Adolesc. 23 (2013) 500–512, http://dx.doi.org/10.1111/jora.12059.

[11] B.R. Hoffman, S. Sussman, J.B. Unger, T.W. Valente, Peer influences on adolescent cigarette smoking: A theoretical review of the literature, Subst. Use Misuse 41 (2006) 103–155, http://dx.doi.org/10.1080/10826080500368892.

[12] J.S. Tucker, K. de la Haye, D.P. Kennedy, H.D. Green, M.S. Pollard, Peer influence on marijuana use in different types of friendships, J. Adolesc. Health 54 (2014) 67–73, http://dx.doi.org/10.1016/j.jadohealth.2013.07.025.

[13] A.D. Reynolds, T.M. Crea, Peer influence processes for youth delinquency and depression, J. Adolesc. 43 (2015) 83–95, http://dx.doi.org/10.1016/j.adolescence.2015.05.013.

[14] S. Wasserman, K. Faust, Social Network Analysis: Methods and Applications, Cambridge University Press, 1994, https://books.google.es/books/about/Social_Network_Analysis.html?id=CAm2DpIqRUIC&redir_esc=y. (Accessed 18 June 2018).

[15] W. De Nooy, A. Mrvar, V. Batagelj, Exploratory Social Network Analysis with Pajek, Cambridge University Press, 2018, http://dx.doi.org/10.1017/9781108565691.

[16] M.A. Brandão, M.M. Moro, Social professional networks: A survey and taxonomy, Comput. Commun. 100 (2017) 20–31, http://dx.doi.org/10.1016/j.comcom.2016.12.011.

[17] J. Kim, M. Hastak, Social network analysis: Characteristics of online social networks after a disaster, Int. J. Inf. Manage. 38 (2018) 86–96, http://dx.doi.org/10.1016/J.IJINFOMGT.2017.08.003.

[18] X. Zenuni, B. Raufi, F. Ismaili, J. Ajdari, State of the art of semantic web for healthcare, Procedia Soc. Behav. Sci. (2015) http://dx.doi.org/10.1016/j.sbspro.2015.06.213.

[19] B. Nie, S. Sun, Knowledge graph embedding via reasoning over entities, relations, and text, Futur. Gener. Comput. Syst. 91 (2019) 426–433, http://dx.doi.org/10.1016/J.FUTURE.2018.09.040.

[20] J.K. Tarus, Z. Niu, A. Yousif, A hybrid knowledge-based recommender system for e-learning based on ontology and sequential pattern mining, Futur. Gener. Comput. Syst. 72 (2017) 37–48, http://dx.doi.org/10.1016/J.FUTURE.2017.02.049.

[21] J. Scott, Social Network Analysis : A Handbook, SAGE Publications, 2000, https://books.google.es/books/about/Social_Network_Analysis.html?id=Ww3_bKcz6kgC&redir_esc=y. (Accessed 18 June 2018).

[22] S. Wasserman, K. Faust, Social Network Analysis: Methods and Applications, 1994, http://dx.doi.org/10.1525/ae.1997.24.1.219.

[23] J. Travers, S. Milgram, An experimental study of the small world problem, Sociometry 32 (1969) 425, http://dx.doi.org/10.2307/2786545.

[24] M.E.J. Newman, Detecting community structure in networks, Eur. Phys. J. B 38 (2004) 321–330, http://dx.doi.org/10.1140/epjb/e2004-00124-y.

[25] J.S. Coleman, Social capital in the creation of human capital, Am. J. Sociol. 94 (1988) http://dx.doi.org/10.2307/2780243.

[26] R.S. Burt, Structural holes and good ideas, Am. J. Sociol. 110 (2004) 349–399, http://dx.doi.org/10.1086/421787.

[27] L.C. Freeman, Centrality in social networks conceptual clarification, Soc. Networks 1 (1978) 215–239, http://dx.doi.org/10.1016/0378-8733(78)90021-7.

[28] U.J. Nieminen, On the centrality in a directed graph, Soc. Sci. Res. 2 (1973) 371–378, http://dx.doi.org/10.1016/0049-089X(73)90010-0.

[29] W.L. Garrison, Connectivity of the interstate highway system, Pap. Reg. Sci. 6 (2005) 121–137, http://dx.doi.org/10.1111/j.1435-5597.1960.tb01707.x.

[30] F.R. Pitts, A graph theoretic approach to historical geography, Prof. Geogr. 17 (1965) 15–20, http://dx.doi.org/10.1111/j.0033-0124.1965.015_m.x.

[31] P. Holme, B.J. Kim, C.N. Yoon, S.K. Han, Attack vulnerability of complex networks, Phys. Rev. E 65 (2002) 056109, http://dx.doi.org/10.1103/PhysRevE.65.056109.

[32] T.W. Valente, Social networks and health, 2010, http://dx.doi.org/10.1093/acprof:oso/9780195301014.001.0001.

[33] D. Chambers, P. Wilson, C. Thompson, M. Harden, Social network analysis in healthcare settings: A systematic scoping review, PLoS One 7 (2012) http://dx.doi.org/10.1371/journal.pone.0041911.

[34] D.L. Haynie, D.A. Kreager, Dangerous liaisons? Dating and drinking diffusion in adolescent peer networks 76, 2011, pp. 737–763, http://dx.doi.org/10.1177/0003122411416934. DANGEROUS.

[35] D.M. Kirke, Chain reactions in adolescents' cigarette, alcohol and drug use: Similarity through peer influence or the patterning of ties in peer networks? Soc. Networks 26 (2004) 3–28, http://dx.doi.org/10.1016/j.socnet.2003.12.001.

[36] A.B. Knecht, W.J. Burk, J. Weesie, C. Steglich, Friendship and alcohol use in early adolescence: A multilevel social network approach, J. Res. Adolesc. 21 (2011) 475–487, http://dx.doi.org/10.1111/j.1532-7795.2010.00685.x.

[37] T.A.B. Snijders, G.G. van de Bunt, C.E.G. Steglich, Introduction to stochastic actor-based models for network dynamics, Soc. Networks 32 (2010) 44–60, http://dx.doi.org/10.1016/j.socnet.2009.02.004.

[38] T.J. Dishion, Stochastic agent-based modeling of influence and selection in adolescence: Current status and future directions in understanding the dynamics of peer contagion, J. Res. Adolesc. 23 (2013) 596–603, http://dx.doi.org/10.1111/jora.12068.

[39] J.M. Light, C.C. Greenan, J.C. Rusby, K.M. Nies, T.A.B. Snijders, Onset to first alcohol use in early adolescence: A network diffusion model, J. Res. Adolesc. (2013) http://dx.doi.org/10.1111/jora.12064.

[40] D.R. Schaefer, A network analysis of factors leading adolescents to befriend substance-using peers, J. Quant. Criminol. 34 (2018) 275–312, http://dx.doi.org/10.1007/s10940-016-9335-4.

[41] G.C. Huang, J.B. Unger, D. Soto, K. Fujimoto, M.A. Pentz, M. Jordan-Marsh, T.W. Valente, Peer influences: The impact of online and offline friendship networks on adolescent smoking and alcohol use, J. Adolesc. Heal. 54 (2014) 508–514, http://dx.doi.org/10.1016/j.jadohealth.2013.07.001.

[42] K. Sanchagrin, K. Heimer, A. Paik, Adolescent. Delinquency, Drinking, Adolescent delinquency drinking and smoking: Does the gender of friends matter? Youth Soc. 49 (2017) 805–826, http://dx.doi.org/10.1177/0044118X14563050.

[43] E. Long, T.S. Barrett, G. Lockhart, Network-behavior dynamics of adolescent friendships, alcohol use, and physical activity, Heal. Psychol. (2017) http://dx.doi.org/10.1037/hea0000483.

[44] K. Fujimoto, T.W. Valente, Multiplex congruity: Friendship networks and perceived popularity as correlates of adolescent alcohol use, Soc. Sci. Med. 125 (2015) 173–181, http://dx.doi.org/10.1016/j.socscimed.2014.05.023.

[45] D.W. Osgood, M.E. Feinberg, L.N. Wallace, J. Moody, Friendship group position and substance use, Addict. Behav. 39 (2014) 923–933, http://dx.doi.org/10.1016/j.addbeh.2013.12.009.

[46] C. Wang, J.R. Hipp, C.T. Butts, R. Jose, C.M. Lakon, Alcohol use among adolescent youth: The role of friendship networks and family factors in multiple school studies, PLoS One 10 (2015) 1–19, http://dx.doi.org/10.1371/journal.pone.0119965.

[47] C. Wang, J.R. Hipp, C.T. Butts, R. Jose, C.M. Lakon, Peer influence, peer selection and adolescent alcohol use: a simulation study using a dynamic network model of friendship ties and alcohol use, Prev. Sci. 18 (2017) 382–393, http://dx.doi.org/10.1007/s11121-017-0773-5.

[48] B. Fitzpatrick, J. Martinez, E. Polidan, E. Angelis, The big impact of small groups on college drinking, J. Artif. Soc. Soc. Simul. 18 (2015) 1–52, http://dx.doi.org/10.18564/jasss.2760.

[49] J.F. Kelly, R.L. Stout, M.C. Greene, V. Slaymaker, Young adults, social networks, and addiction recovery: Post treatment changes in social ties and their role as a mediator of 12-step participation, PLoS One (2014) http://dx.doi.org/10.1371/journal.pone.0100121.

[50] F.F. Ishtaiwa, I.M. Aburezeq, The impact of Google Docs on student collaboration: A UAE case study, Learn. Cult. Soc. Interact. 7 (2015) 85–96, http://dx.doi.org/10.1016/j.lcsi.2015.07.004.

[51] A. Gordon, SurveyMonkey.com—Web-based survey and evaluation system: , Internet High. Educ. 5 (2002) 83–87, http://dx.doi.org/10.1016/S1096-7516(02)00061-1, http://www.surveymonkey.com.

[52] Limesurvey GmbH, Limesurvey: An Open Source Survey Tool, LimeSurvey GmbH, Hamburg, Ger., 2017.

[53] M.J. Bohn, T.F. Babor, H.R. Kranzler, The alcohol use disorders identification test (AUDIT): validation of a screening instrument for use in medical settings, J. Stud. Alcohol. 56 (1995) 423–432, http://www.ncbi.nlm.nih.gov/pubmed/7674678. (Accessed 18 June 2018).

[54] S.P. Borgatti, M.G. Everett, L.C. Freeman, Ucinet for windows: Software for social network analysis, Harvard Anal. Technol. (2002) http://dx.doi.org/10.1111/j.1439-0310.2009.01613.x.

[55] L. Liu, M.T. Özsu, Encyclopedia of Database Systems, Springer, 2009.

[56] T. Berners-Lee, J. Hendler, O. Lassila, The semantic web, Sci. Am. (2001) http://dx.doi.org/10.1038/scientificamerican0501-34.

[57] D. McGuinness, F. van Harmelen, OWL web ontology language overview, 2009, https://www.w3.org/TR/owl-features/. (Accessed 18 June 2018).

[58] K.S. Candan, H. Liu, R. Suvarna, Resource description framework: Metadata and its applications, SIGKDD Explor. (2001) http://dx.doi.org/10.1145/507533.507536.

[59] E. Prud'hommeaux, A. Seaborne, SPARQL Query Language for RDF, W3C Recomm., 2008.

[60] I. Horrocks, P.F. Patel-schneider, H. Boley, S. Tabet, B. Grosof, M. Dean, SWRL : A Semantic Web Rule Language Combining OWL and RuleML, W3C Memb. Submiss. 21, 2004.

[61] G. Erétéo, F. Limpens, F. Gandon, O. Corby, M. Buffa, M. Leitzelman, P. Sander, Semantic social network analysis, in: Handb. Res. Methods Tech. Stud. Virtual Communities, IGI Global, 2011, pp. 122–156, http://dx.doi.org/10.4018/978-1-60960-040-2.ch007.

[62] J.C. Paolillo, E. Wright, Social network analysis on the semantic web: Techniques and challenges for visualizing FOAF, Vis. Semant. Web XML-Based Internet Inf. Vis. (2006) 229–241, http://dx.doi.org/10.1007/1-84628-290-X_14.

[63] P. Mika, Social Networks and the Semantic Web, first ed., Springer, Barcelona, 2007.

[64] J. Golbeck, M. Rothstein, Linking social networks on the web with FOAF, Soc. Networks (2008).

[65] S.Y. Bhat, M. Abulaish, Analysis and mining of online social networks: Emerging trends and challenges, Wiley Interdiscip. Rev. Data Min. Knowl. Discov. (2013) http://dx.doi.org/10.1002/widm.1105.

[66] C.E. Currie, R.A. Elton, J. Todd, S. Platt, Indicators of socioeconomic status for adolescents: the WHO health behaviour in school-aged children survey, Health Educ. Res. 12 (1997) 385–397, http://www.ncbi.nlm.nih.gov/pubmed/10174221. (Accessed 18 June 2018).

[67] U. Ravens-Sieberer, A. Gosch, M. Erhart, U. von Rueden, The Kidscreen Questionnaires: Handbook, 2006.

[68] D. Brickley, L. Miller, FOAF Vocabulary specification 0.99, 2000, http://xmlns.com/foaf/spec/. (Accessed 16 August 2017).

[69] M. Alipour-Aghdam, Ontology-driven generic questionnaire design, 2014, p. 85.

[70] G. Ereteo, Semantic social network analysis, Telecom ParisTech, 2011, https://tel.archives-ouvertes.fr/tel-00586677.

[71] E.M. Beniaminov, Ontology libraries on the web: Status and prospects, Autom. Doc. Math. Linguist. 52 (2018) 117–120, http://dx.doi.org/10.3103/s0005105518030020.

[72] J.J. Carroll, C. Bizer, P. Hayes, P. Stickler, Named graphs, provenance and trust, in: Proc. 14th Int. Conf. World Wide Web, WWW '05, ACM Press, New York, New York, USA, 2005, p. 613, http://dx.doi.org/10.1145/1060745.1060835.

[73] E. Sirin, B. Parsia, B.C. Grau, A. Kalyanpur, Y. Katz, Pellet: A practical OWL-DL reasoner, Web Semant. 5 (2007) 51–53, http://dx.doi.org/10.1016/j.websem.2007.03.004.

[74] S.P. Borgatti, M.G. Everett, L.C. Freeman, Ucinet IV: Network analysis software, 1992.

[75] A. Sapountzi, K.E. Psannis, Social networking data analysis tools & challenges, Futur. Gener. Comput. Syst. 86 (2018) 893–913, http://dx.doi.org/10.1016/j.future.2016.10.019.

[76] N. Akhtar, Social network analysis tools, in: Proc. 2014 Fourth Int. Conf. Commun. Syst. Netw. Technol., IEEE Computer Society, Washington, DC, USA, 2014, pp. 388–392, http://dx.doi.org/10.1109/CSNT.2014.83.

[77] K. Cherven, Network Graph Analysis and Visualization with Gephi, Packt Publishing Ltd, 2013.

[78] T.W. Valente, Network interventions, Science (80-.) 336 (2012) 49–53, http://dx.doi.org/10.1126/science.1217330.

[79] S. Cavallari, V.W. Zheng, H. Cai, K.C.C. Chang, E. Cambria, Learning community embedding with community detection and node embedding on graphs, in: Int. Conf. Inf. Knowl. Manag. Proc., Association for Computing Machinery, 2017, pp. 377–386, http://dx.doi.org/10.1145/3132847.3132925.

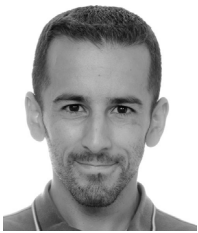

**José Alberto Benítez-Andrades, Ph.D.** was born in Granada, Spain, in 1988. He has received his degree in Computer Science from the University of León, and the Ph.D. degree in Production and Computer Engineering in 2017 (University of Leon). He was part time instructor who kept a parallel job from 2013 to 2018 and since 2018 he works as teaching assistant at the University of Leon. His research interests include artificial intelligence, knowledge engineering, semantic technologies. He was a recipient of award to the Best Doctoral Thesis 2018 by Colegio Profesional de Ingenieros en Informática en Castilla y León in 2018.

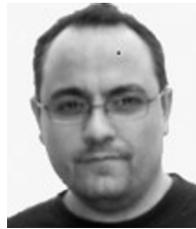

**Isaías García-Rodríguez, Ph.D.** received his Bachelor degree in Industrial Technical Engineering from the University of León (Spain) in 1992 and her Master degree in Industrial Engineering from the University of Oviedo (Spain) in 1996. Isaías obtained his Ph.D. in Computer Science from the University of León in 2008, where he is currently a lecturer. His current research interests include practical applications of Software Defined Networks, Network Security and applied Knowledge Engineering techniques. He has published different scientific papers in journals, Conferences and Symposia around the world.

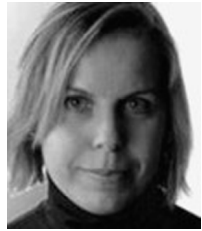

**Carmen Benavides, Ph.D.** received her Bachelor degree in Industrial Technical Engineer from the University of León (Spain) in 1996 and her Master degree in Electronic Engineering from the University of Valladolid (Spain) in 1998. Carmen obtained her Ph.D. in Computer Science from the University of León in 2009 and she works as an Assistant Professor at the same University since 2001. Her research interests are focused on applied Knowledge Engineering techniques, practical applications of Software Defined Networks and Network Security. She has organized several congresses, and has presented and published different papers in Journals, Conferences and Symposia.

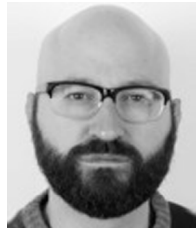

**Héctor Alaiz-Moretón, Ph.D.** received his degree in Computer Science, performing the final project at Dublin Institute of Technology, in 2003. He received his Ph.D. in Information Technologies in 2008 (University of Leon). He has worked like a lecturer since 2005 at the School of Engineering at the University of Leon. His research interests include knowledge engineering, machine & deep learning, networks communication and security. He has several works published in international conferences, as well as books and scientific papers in peer review journals. He has been member of scientific committees in conferences. He has headed several Ph.D. Thesis and research projects.

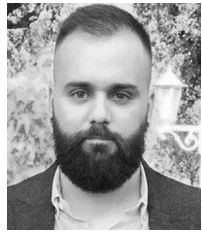

**Alejandro Rodríguez-González, Ph.D.**, is an Associate Professor at the Department of Computer Languages and Systems and Software Engineering at Universidad Politécnica de Madrid and the principal investigator of the Medical Data Analysis laboratory (MEDAL) at Center for Biomedical Technology. His main research interests are Artificial Intelligence and Biomedical informatics field, with an interest on knowledge representation and extraction. Prof. Rodríguez was awarded in January 2018 with the "Best UPM Research Trajectory" award (2017 edition). Prof. Rodríguez is involved in several H2020 projects about Data Science and Big data in different domains, including skills definition, health sector and cyber physical products.